\documentclass[10pt,twocolumn,letterpaper]{article}

\usepackage[pagenumbers]{cvpr} % To force page numbers, e.g. for an arXiv version

\usepackage{graphicx}
\usepackage{amsmath}
\usepackage{amssymb}
\usepackage{booktabs}

\usepackage{comment}
\usepackage{color}
\usepackage{wrapfig}

\usepackage{booktabs}
\usepackage{algorithm}
\usepackage{algorithmic}
\usepackage{multirow}
\usepackage{tabularx}
\usepackage{verbatim}
\usepackage{subcaption}
\usepackage{pifont}
\usepackage{xcolor}

\newcommand{\gr}[1]{\textcolor{gray}{#1}}

\usepackage[pagebackref=true,breaklinks=true,colorlinks,bookmarks=false]{hyperref}
\usepackage[capitalize]{cleveref}
\crefname{section}{Sec.}{Secs.}
\Crefname{section}{Section}{Sections}
\Crefname{table}{Table}{Tables}
\crefname{table}{Tab.}{Tabs.}

\def\cvprPaperID{*****} % *** Enter the CVPR Paper ID here
\def\confName{CVPR}
\def\confYear{2022}

\begin{document}

\title{Matting Anything}
% \author{Jiachen Li\textsuperscript{1}
% \hspace{10mm}
% Jitesh Jain\textsuperscript{1,2}
% \hspace{10mm}
% Humphrey Shi\textsuperscript{1,3} \\
% {\small \textsuperscript{1}SHI Labs $@$ University of Oregon \& UIUC
% \hspace{5mm}
% \textsuperscript{2}IIT Roorkee
% \hspace{5mm}
% \textsuperscript{3}Picsart AI Research (PAIR)}\\
% {\small \url{https://chrisjuniorli.github.io/project/Matting-Anything/} 
% }
% }

\author{
    Jiachen Li\textsuperscript{1},
    Jitesh Jain\textsuperscript{1},
    Humphrey Shi\textsuperscript{1,2} \\
{\small \textsuperscript{1}SHI Labs $@$ Georgia Tech \& Oregon \& UIUC,
% \textsuperscript{5}Google Research,
\textsuperscript{2}Picsart AI Research (PAIR)}\\
{\small \textbf{\url{https://github.com/SHI-Labs/Matting-Anything}}}
}

%\maketitle

\twocolumn[{
\maketitle
\begin{center}
    \centering
    \captionsetup{type=figure}
    \includegraphics[width=1.0\textwidth]{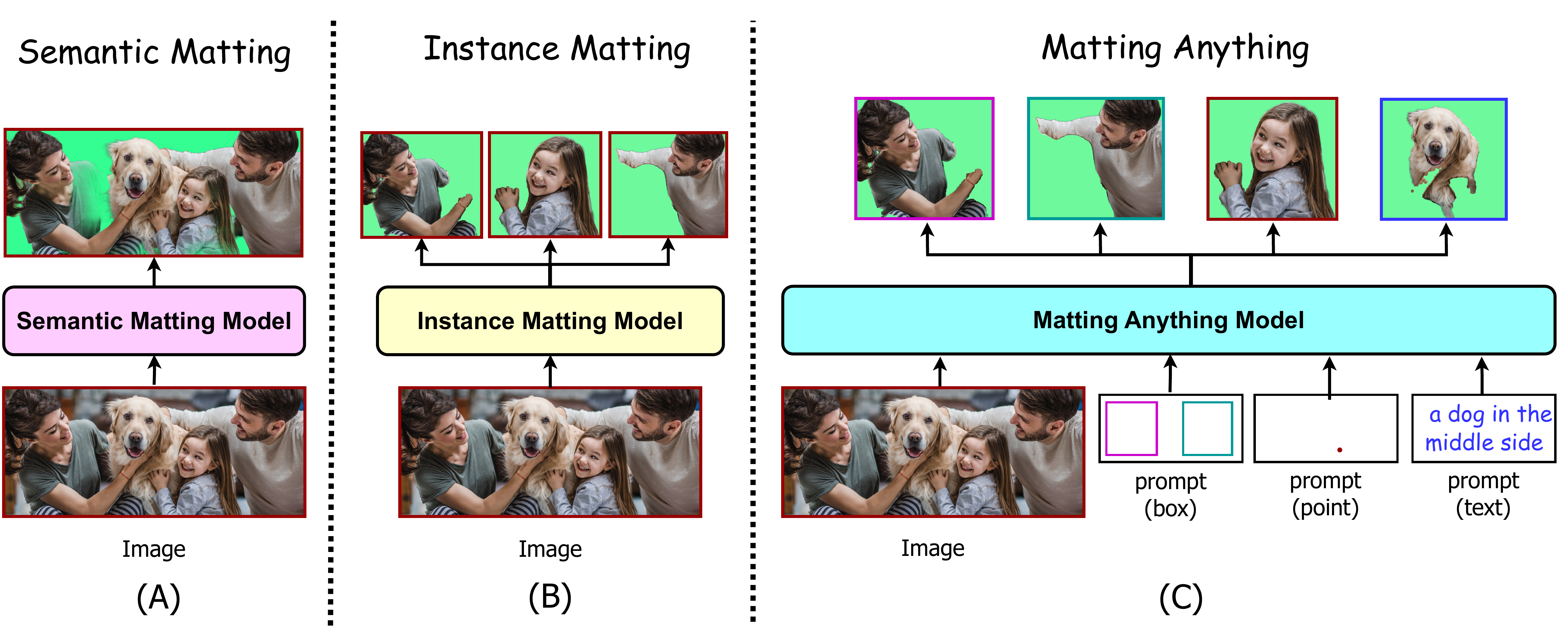}
    \captionof{figure}{\textbf{Matting Anything Model}~(MAM) offers a versatile framework capable of addressing various types of image matting scenarios with a single model. Compared to previous specialized models for (A) Semantic Matting, which outputs a single alpha matte of all instances in the foreground; (B) Instance Matting, which returns alpha mattes of all human instances;
    (C) Matting Anything Model can estimate the alpha matte of any target instance with user prompts as boxes, points, or text descriptions for interactive use by incorporating SAM~\cite{kirillov2023segment}. It further reaches comparable performance to the specialized matting models on multiple benchmarks, and shows superior generalization ability with fewer parameters as a unified image matting model.}
    \label{fig:teaser}
\end{center}
}]

\begin{abstract}
     In this paper, we propose the Matting Anything Model (MAM), an efficient and versatile framework for estimating the alpha matte of any instance in an image with flexible and interactive visual or linguistic user prompt guidance. MAM offers several significant advantages over previous specialized image matting networks: (i) MAM is capable of dealing with various types of image matting, including semantic, instance, and referring image matting with only a single model; (ii) MAM leverages the feature maps from the Segment Anything Model (SAM)~\cite{kirillov2023segment} and adopts a lightweight Mask-to-Matte (M2M) module to predict the alpha matte through iterative refinement, which has only 2.7 million trainable parameters. (iii) By incorporating SAM, MAM simplifies the user intervention required for the interactive use of image matting from the trimap to the box, point, or text prompt. We evaluate the performance of MAM on various image matting benchmarks, and the experimental results demonstrate that MAM achieves comparable performance to the state-of-the-art specialized image matting models under different metrics on each benchmark. Overall, MAM shows superior generalization ability and can effectively handle various image matting tasks with fewer parameters, making it a practical solution for unified image matting.
     Our code and models are open-sourced at \href{https://github.com/SHI-Labs/Matting-Anything}{https://github.com/SHI-Labs/Matting-Anything}.

\end{abstract}

%%%%%%%%% BODY TEXT
\section{Introduction}

Image Matting, as a long-standing computer vision task, aims to estimate the alpha matte $\alpha$ given an input image $I$~\cite{wang2008image}. The matting target is mainly around human beings or other objects at the semantic level~\cite{li2021deep, sun2021semantic, xu2022situational}. Recent works have extended the scope of image matting to more complex scenarios like image instance matting~\cite{sun2022human}, which requires instance-aware alpha matte predictions and referring image matting~\cite{li2023referring}, which extracts the alpha matte given natural language description. 

Previous deep learning-based image matting methods~\cite{xu2017deep,wang2018deep,zhu2017fast,li2020natural,yu2020high,qiao2020attention, Park_2022_CVPR, sun2022human, li2023referring} have been proposed to address specific image matting tasks on corresponding benchmarks. These methods are tailored to individual datasets and lack the flexibility to handle various image matting tasks due to their fixed model designs. This limitation has hindered the development of more generalized and versatile image matting models. As a result, there is a growing interest in developing more adaptive and efficient image matting frameworks that can handle different types of image matting tasks with a single model.

Furthermore, previous image matting methods have relied on user-guided trimaps as auxiliary inputs to achieve accurate alpha matte predictions. Although some trimap-free methods have been proposed that use mask guidance or background images instead~\cite{yu2021mask, sengupta2020background}, they are unable to estimate the alpha matte of the target instance based on the user request for interactive use. Therefore, it is crucial to develop a model that can achieve accurate alpha matte estimation without relying on user-guided trimaps, while also being capable of handling simple user requests in a flexible and efficient manner for interactive use. Such a model would significantly enhance the user experience by reducing the extra need for manual intervention.

Motivated by these limitations of image matting, we propose the Matting Anything Model (MAM), a versatile network that can estimate the alpha matte of any target instance with prompt-based user guidance in an image as shown in Figure~\ref{fig:teaser}. MAM leverages the recent Segment Anything Model (SAM) framework~\cite{kirillov2023segment}, which supports flexible prompting and outputs segmentation masks of any target instance for interactive use. Specifically, MAM takes the feature maps and mask outputs from SAM as inputs and adds a lightweight Mask-to-Matte (M2M) module to predict the alpha matte of the target instance. We trained MAM on a combination of five image matting datasets that cover different classes of instances, allowing the M2M module to learn generalizable features for image matting. During training, we randomly place target instances onto background images and use a pre-trained SAM to output mask predictions of the corresponding instances. The trainable M2M module then refines the mask by predicting multi-scale alpha mattes. Through an iterative refinement process based on the mask or the alpha matte, the multi-scale predictions are merged to obtain the final meticulous alpha matte.

We conducted extensive evaluations of MAM on six image matting benchmarks, including semantic image matting benchmark PPM-100~\cite{sun2021modnet}, AM2K~\cite{li2022bridging} PM-10K~\cite{li2022bridging}, the instance image matting benchmark RWP636~\cite{yu2021mask}, HIM2K~\cite{sun2022human}, and the referring image matting benchmark RefMatte-RW100~\cite{li2023referring}. Our results demonstrate that MAM achieves performance comparable to that of state-of-the-art image matting models across all benchmarks under different evaluation metrics. The experimental results highlight the versatility and effectiveness of our proposed approach for handling various image matting tasks in an interactive and efficient manner.

\begin{figure*}[tb]
\centering
\includegraphics[width=1.0\textwidth]{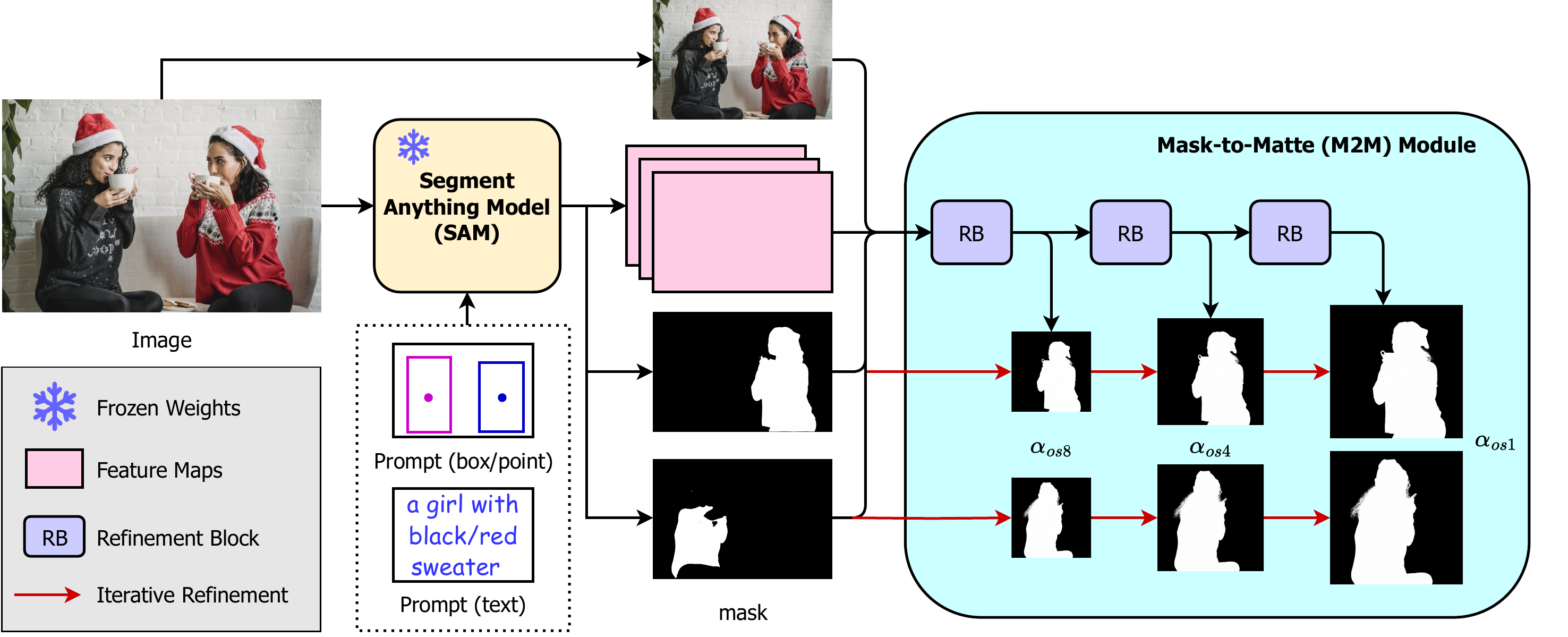}
\caption{\textbf{Matting Anything Model Architecture}. The MAM architecture consists of a pre-trained SAM and an M2M module. Given an input image $I$, SAM generates the mask prediction for the target instance based on the box or point user prompt. The M2M module takes the concatenated inputs, including the image, mask, and feature maps, and produces multi-scale predictions $\alpha_{os8}$, $\alpha_{os4}$, and $\alpha_{os1}$. The iterative refinement process, detailed in Section~\ref{method}, progressively improves the precision of the final meticulous alpha matte $\alpha$, incorporating information from the multi-scale outputs.}
\label{fig:archi}
\vspace{-3mm}
\end{figure*}

\section{Related Works}
\subsection{Image Matting} 
Given an image $I$, which can be view as a combination of foreground image $F$ and background image $B$ with coefficient alpha matte $\alpha$, 
\begin{equation}
    I = \alpha F + (1 - \alpha) B
\end{equation}
Image Matting is to estimate $\alpha$ given only $I$ as inputs. Traditional methods rely on a user-guided trimap, which explicitly annotates the absolute foreground area, absolute background area, and transition area. Then, sampling-based image matting solutions use low-level features to distinguish the transition areas by measuring the similarities between foreground and background neighbors~\cite{aksoy2017designing,chuang2001bayesian,bai2007geodesic,chen2013knn,grady2005random,feng2016cluster}.
Recently, deep learning-based methods~\cite{xu2017deep,wang2018deep,zhu2017fast,yu2020high, Park_2022_CVPR, li2023deep} adopt neural networks to estimate the alpha matte in an end-to-end manner with trimap as auxiliary inputs. Some trimap-free methods use background image~\cite{sengupta2020background}, mask guidance~\cite{liu2020boosting, yu2021mask}, or segmentation data~\cite{chen2018semantic, lin2021robust} to make up the absence of trimap. When the image $I$ contains multiple instances, the composition turns to
\begin{equation}
    I = \sum_i^N \alpha_i F_i + (1 - \sum_i^N \alpha_i) B
\end{equation}
$\alpha_i$ represents the alpha matte of instance $i$ and InstMatt~\cite{sun2022human} adopt the target and reference mask as guidance to prediction instance-aware alpha matte prediction. Interactive matting methods~\cite{yang2022unified, wei2021improved, li2023referring} develop specialized models that use point, boxes, or text input to estimate the alpha matte of the target instance. MatAny~\cite{yao2023matte} is a concurrent work that also adopts SAM for semantic image matting. In terms of video matting, trimap-free~\cite{sun2021modnet, lin2021robust, li2023videomatt, li2022vmformer, li2023video} methods are explored for real-time inference while the per-frame prediction quality is not comparable to image matting methods. However, these methods are designed for a certain scenario with corresponding benchmarks, which limits their potential to handle various image matting tasks and benchmarks simultaneously.

\subsection{Image Segmentation}
Image segmentation is a close research area to image matting, while it predicts the binary mask of different instances in the image. Similar to image matting, many image segmentation methods are tailored for a specific image segmentation task, like semantic segmentation~\cite{huang2019ccnet, chen2017rethinking}, instance segmentation~\cite{he2017mask, wang2020solov2}, and panoptic segmentation~\cite{wang2021max, kirillov2019panoptic}. Recent works started to explore transformer-based frameworks~\cite{cheng2021masked, jain2021semask, jain2022oneformer, hassani2023neighborhood} for unified image segmentation. Language-guided segmentation frameworks~\cite{xu2022groupvit, zhou2023lmseg} look for text supervision to segment instance-aware masks.
OneFormer~\cite{jain2022oneformer} adopts a single transformer model to learn with a joint training strategy and performs universal segmentation across semantic, instance and panoptic segmentation and outperforms specialized models.
SAM~\cite{kirillov2023segment} takes a further step recently, which supports flexible prompting from users to segment any instance in an image for interactive use. Grounded-SAM~\cite{liu2023grounding} incorporates DINO with SAM to add text prompt support. Foundation models like SAM offer opportunities for other areas to develop versatile frameworks to support a range of applications. 
\section{Matting Anything}
\label{method}

In this section, we provide an overview of the Matting Anything Model (MAM) architecture, which consists of two main components: the frozen Segment Anything Model (SAM) and the trainable Mask-to-Matte (M2M) module. We first provide a brief review of the SAM, which is designed to produce high-quality instance segmentation given user-guided prompts. We then introduce the M2M module, which enables the transformation of the binary masks into high-quality alpha mattes. Finally, we describe how we connect the M2M module with the SAM to gradually build the end-to-end MAM.

\begin{table*}[tb]
\centering
\resizebox{1.0\textwidth}{!}{
\begin{tabular}{c|c|c|c|cc|cc|cc}
Task &\multicolumn{3}{c|}{Semantic Matting}  &\multicolumn{4}{c|}{Instance Matting}
 &\multicolumn{2}{c}{Referring Matting}\\ \hline
Benchmark &AM2K &PM-10K &PPM-100 &\multicolumn{2}{c|}{HIM2K} &\multicolumn{2}{c|}{RWP636} &\multicolumn{2}{c}{RefMatte-RW100} \\ \hline
Metric   
&SAD$_{all}$$\downarrow$ 
&MAD$_{all}$$\downarrow$
&MSE$_{all}$$\downarrow$  &IMQ$^{nat}_{mad}$$\uparrow$ &IMQ$^{nat}_{mse}$$\uparrow$ &IMQ$_{mad}$$\uparrow$ 
&IMQ$_{mse}$$\uparrow$ 
&SAD$_{all}$$\downarrow$
&MSE$_{all}$$\downarrow$\\ \hline \hline
\multicolumn{2}{l}{\textit{\gr{Specialized Models}}} \\ \hline \hline
\gr{GFM-R}~\cite{li2022bridging} &\gr{10.89} &\gr{6.7} &\gr{-} &\gr{-} &\gr{-} &\gr{-}  &\gr{-} &\gr{-} &\gr{-}   \\
\gr{GFM-D}~\cite{li2022bridging} &\gr{10.26} &\gr{6.9} &\gr{-} &\gr{-} &\gr{-} &\gr{-}  &\gr{-} &\gr{-} &\gr{-}   \\
\gr{MODNet}~\cite{ke2022modnet} &\gr{-} &\gr{-} &\gr{4.4} &\gr{-} &\gr{-} &\gr{-}  &\gr{-} &\gr{-} &\gr{-}   \\
\gr{MGMatting}~\cite{yu2021mask} &\gr{-} &\gr{-} &\gr{-} &\gr{57.98} &\gr{71.12} &\gr{30.64} &\gr{53.16} &\gr{-} &\gr{-} \\
%\gr{InstMatt}~\ref{} &\gr{-} &\gr{-} &\gr{-} &\gr{70.26} &\gr{81.34} &\gr{51.10} &\gr{73.09} &\gr{-} &\gr{-} \\
\gr{InstMatt}~\cite{sun2022human} &\gr{-} &\gr{-} &\gr{-} &\gr{70.26} &\gr{81.34} &\gr{51.10} &\gr{73.09} &\gr{-} &\gr{-} \\
\gr{CLIPMat-B}~\cite{li2023referring} &\gr{-} &\gr{-} &\gr{-} &\gr{-} &\gr{-} &\gr{-} &\gr{-} &\gr{107.81} &\gr{59.5} \\
\gr{CLIPMat-L}~\cite{li2023referring} &\gr{-} &\gr{-} &\gr{-} &\gr{-} &\gr{-} &\gr{-} &\gr{-} &\gr{85.83} &\gr{47.4} \\
\hline \hline
\multicolumn{2}{l}{\textit{Generalized Models}} \\ \hline \hline
SAM~\cite{kirillov2023segment} &25.00 &25.7  &10.8 &61.15 &74.01  &49.87 &56.92 &33.51 &17.9 \\ 
MAM  &17.30 &15.4 &4.6 &68.78 &81.67  &54.40 &76.45 &29.24 &15.1 \\
\end{tabular}}
\caption{Comparisons between specialized matting models and MAM on various benchmarks. $\uparrow$ / $\downarrow$ means higher / lower values indicate better performance for the corresponding metric. \gr{Gray} text refers to models specifically designed for these benchmarks. MAM shows clear improvements over SAM and superior generalization ability as a unified image matting model.} 
\vspace{-3mm}
\label{tab:specialvsmam}
\end{table*}

\subsection{Segment Anything Model}
\label{sam}
Segment Anything is a recently proposed foundation model for segmentation. Given an image $I \in \mathbb{R}^{3 \times H \times W}$, SAM uses a ViT-based image encoder to obtain deep feature maps $F \in \mathbb{R}^{C \times \frac{H}{16} \times \frac{W}{16}}$. Then, a variety of $N$ input prompts are encoded by the prompt encoder and sent to the mask decoder with the feature maps. The mask decoder returns a set of mask candidates $m_i \in \mathbb{R}^{1 \times H \times W}, i \in N$ indicated by the input prompts. With its flexible prompting mechanism, SAM allows for interactive use and is easily adaptable for downstream tasks.

\subsection{Mask-to-Matte}
\label{m2m}
The Mask-to-Matte (M2M) module is an integral component of our Matting Anything Model (MAM) and is designed to convert instance-aware mask predictions from SAM into instance-aware alpha matte predictions efficiently and smoothly. To achieve this, we utilize the feature maps and mask predictions generated by SAM as auxiliary inputs to M2M. To improve the accuracy of our predictions, we adopt multi-scale branches for predicting the alpha matte and merge these predictions through an iterative refinement schedule.

\noindent \textbf{Multi-Scale Prediction:} Given an input image $I \in \mathbb{R}^{3 \times H \times W}$, the pre-trained SAM model produces feature maps $F \in \mathbb{R}^{C \times \frac{H}{16} \times \frac{W}{16}}$ and mask prediction $m \in \mathbb{R}^{1 \times H \times W}$ on the target instance with prompt guidance. We concatenate the rescaled image, mask, and feature maps to form the input $F_{m2m} \in \mathbb{R}^{(C+4) \times \frac{H}{8} \times \frac{W}{8}}$ to the M2M module. M2M employs several refinement blocks~\cite{cheng2020cascadepsp, yu2021mask}, which contain connected self-attention layer~\cite{zhang2019self}, batchnorm layer, and activation layer, to generate alpha matte predictions at $1/8$ resolution, denoted as $\alpha_{os8} \in \mathbb{R}^{1 \times \frac{H}{8} \times \frac{W}{8}}$. The feature maps are then upsampled to higher resolutions to make alpha matte predictions at $1/4$ and full resolution, denoted as $\alpha_{os4} \in \mathbb{R}^{1 \times \frac{H}{4} \times \frac{W}{4}}$ and $\alpha_{os1} \in \mathbb{R}^{1 \times H \times W}$, respectively. The multi-scale predictions enable MAM to handle objects of varying scales and provide finer-grained alpha mattes for detailed object extraction.

\noindent \textbf{Iterative Refinement} To improve the accuracy of global and local predictions, we use an iterative refinement process. We first compute weight maps $w_{os8}$, $w_{os4}$, and $w_{os1}$ that highlight different areas of the image during training like trimaps. These weight maps are used to compute losses for each scale of prediction, with $w_{os8}$ emphasizing the entire image for $\alpha_{os8}$ predictions, $w_{os4}$ filtering out the background for $\alpha_{os4}$ predictions, and $w_{os1}$ focusing only on the transition areas. During inference, we gradually merge the predictions of $\alpha_{os8}$, $\alpha_{os4}$, and $\alpha_{os1}$ with the mask predictions $m$ from SAM to obtain the final alpha matte prediction $\alpha \in \mathbb{R}^{1 \times H \times W}$.

\subsection{Matting Anything Model}
\label{mam}
After the development of the Mask-to-Matte (M2M) module, we integrate it with the Segment Anything Model (SAM) to enable end-to-end training and inference for the Matting Anything Model (MAM). This integration allows for a comprehensive and unified framework that handles the entire matting process, from feature extraction to alpha matte prediction. 

\begin{table*}[tb]
\centering
%\resizebox{1.0\textwidth}{!}{
\begin{tabular}{c|ccccc}
 &\multicolumn{5}{c}{AM2K / PM-10K} \\
Method  &SAD$_{all}$$\downarrow$ &MSE$_{all}$$\downarrow$  &MAD$_{all}$$\downarrow$ &Grad$_{all}$$\downarrow$ &SAD$_{tri}$$\downarrow$ \\ \hline
SHM~\cite{chen2018semantic} &17.81 / 16.64 &6.8 / 6.9 &10.2 / 9.7 &12.54 / 14.54  &10.26 / 8.53 \\
LFM~\cite{zhang2019late}  &36.12 / 37.51  &11.6 / 15.2 &21.0 / 15.2 &21.06 / 21.82  &19.68 / 16.36  \\ 
HATT~\cite{qiao2020attention}  &28.01 / 22.66 &5.5 / 3.8 & 16.1 / 13.1 &18.29 / 15.16   &13.36 / 9.32 \\ 
SHMC~\cite{liu2020boosting} &61.50 / 57.85 &27.0 / 29.1 &35.6 / 34.0 &37.00 / 37.28  &35.23 / 23.04 \\ 
\gr{GFM-R}~\cite{li2022bridging} &\gr{10.89 / 11.52} &\gr{2.9 / 3.8} &\gr{6.4 / 6.7} &\gr{10.00 / 13.07}  &\gr{9.15 / 8.00}  \\
\gr{GFM-D}~\cite{li2022bridging} &\gr{10.26 / 11.89}  &\gr{2.9 / 4.1} &\gr{5.9 / 6.9} &\gr{8.82 / 12.90}  &\gr{8.24 / 7.80} \\
\hline
SAM~\cite{kirillov2023segment} &25.00 / 44.11  & 10.8 / 28.8 &14.8 / 25.7 &60.01 / 24.56 &20.72 / 31.96  \\
MAM &17.30 / 25.82 &3.5 / 9.2 &10.1 / 15.4 &10.65 /14.22  &15.67 / 23.99\\
\hline
\end{tabular}
\caption{Results on the semantic image matting benchmark AM2K and PM-10K. Metrics with \textit{all} and \textit{tri} as subscript indicates the evaluation of the whole image and the transition area, separately. $\downarrow$ means lower values indicate better performance for the metric. } 
\label{tab:am2k_pm10k}
\vspace{-2mm}
\end{table*}
\begin{table}[tb]
\centering
\begin{tabular}{c|cc}
Method &MSE$_{all}$ $\downarrow$ &MAD$_{all}$ $\downarrow$ \\ \hline
DIM~\cite{xu2017deep} &11.5 &17.8 \\
FDMPA~\cite{zhu2017fast} &10.1 &16.0 \\
LFM~\cite{zhang2019late} &9.4 &15.8 \\
SHM~\cite{chen2018semantic} &7.2 &15.2 \\
HATT~\cite{qiao2020attention} &6.7 &13.7 \\
BSHM~\cite{liu2020boosting} &6.3 &11.4 \\
\gr{MODNet}~\cite{ke2022modnet} &\gr{4.4} &\gr{8.6} \\
\hline
SAM~\cite{kirillov2023segment} &10.8  &13.8  \\
MAM &4.6  &9.9  \\ \hline
%% SAM Vit-l &9.0 & 12.0
%% MAM Vit-l &2.5 &7.2
%% SAM Vit-H &9.4 &12.3
%% MAM Vit-H & & 
\end{tabular}
\caption{Results on the semantic image matting benchmark PPM-100.}
\vspace{-7mm}
\label{tab:ppm}
\end{table}

\noindent \textbf{Multi-Dataset Training} To ensure the robustness and versatility of our Matting Anything Model (MAM), we adopt a multi-dataset training approach that encompasses diverse foreground instances and background images from various image matting datasets. This selection allows us to cover a wide range of instance classes and background scenarios, enhancing the model's ability to handle different types of instances and backgrounds effectively.
During the training process, we create composite images by combining a foreground instance $F \in \mathbb{R}^{3 \times H \times W}$ with its corresponding ground truth alpha matte $\alpha_{gt} \in \mathbb{R}^{1 \times H \times W}$ and a background image $B \in \mathbb{R}^{3 \times H \times W}$. The composition is performed using the equation $I = \alpha_{gt} F + (1 - \alpha_{gt}) B$. We then extract the bounding box $(x_0, y_0, x_1, y_1)$ that encapsulates the instance of interest within the composite image. 
Then, we send the image $I$ and the bounding box as a prompt to the pre-trained SAM, which returns the mask prediction of the instance. Then, we concatenate the image, mask and feature maps, and send them to the M2M module, which further returns the multi-scale alpha matte predictions $\alpha_{os8}, \alpha_{os4}, \alpha_{os1}$. The loss $\mathcal{L}$ is computed between the multi-scale predictions and ground truth $\alpha_{gt}$ as
\begin{equation}
    \mathcal{L}(\alpha_{gt}, \alpha_{os1}, \alpha_{os4}, \alpha_{os8}) =  \lambda_{\mathcal{L}_1} \mathcal{L}_1 + \lambda_{\mathcal{L}_{Lap}}\mathcal{L}_{Lap}
\end{equation}
$\mathcal{L}_1$ is L1 loss and $\mathcal{L}_{Lap}$ is Laplacian loss used in ~\cite{lin2021robust,hou2019context,sun2021deep}. The coefficients $\lambda_{\mathcal{L}1}$ and $\lambda{\mathcal{L}_{Lap}}$ control the contribution of each loss term, respectively. Both loss terms are computed on multi-scale predictions as
\begin{equation}
    \mathcal{L}_1  = \mathcal{L}_1(\alpha_{gt}, \alpha_{os1})+\mathcal{L}_1(\alpha_{gt}, \alpha_{os4})+\mathcal{L}_1(\alpha_{gt}, \alpha_{os8})
\end{equation}
\begin{equation}
    \mathcal{L}_{Lap}  = \mathcal{L}_{Lap}(\alpha_{gt}, \alpha_{os1})+\mathcal{L}_{Lap}(\alpha_{gt}, \alpha_{os4})+\mathcal{L}_{Lap}(\alpha_{gt}, \alpha_{os8})
\end{equation}

\noindent \textbf{Multi-Benchmark Inference} 
During the inference phase, we conducted extensive evaluations of the Matting Anything Model (MAM) on multiple image matting benchmarks to assess its generality and adaptability.
Given an input image $I$, SAM produced the initial mask prediction $m \in \mathbb{R}^{1 \times H \times W}$, which captured the rough delineation of the instance. Subsequently, M2M contributed to the refinement of the alpha matte prediction by providing multi-scale predictions $\alpha_{os8}$, $\alpha_{os4}$, and $\alpha_{os1}$.
Then, following the iterative refinements, we progressively updated the predictions by replacing the corresponding regions in the mask prediction $m$ with the respective multi-scale predictions that demonstrated positive weight maps, while in some simple cases the replacement is directly done upon $\alpha_{os8}$ instead of $m$. This iterative refinement allowed us to refine the alpha matte estimation iteratively and enhance the precision of the final prediction $\alpha \in \mathbb{R}^{1 \times H \times W}$.
\section{Experiments}
\begin{table*}[tb]
\centering
\begin{tabular}{c|c|cccc|cccc}
&Model &\multicolumn{4}{c}{Synthetic Subset $\uparrow$}  &\multicolumn{4}{c}{Natural Subset $\uparrow$} \\
Method  & Size   &IMQ$_{mad}$ &IMQ$_{mse}$ &IMQ$_{grad}$   &IMQ$_{conn}$ &IMQ$_{mad}$ &IMQ$_{mse}$ &IMQ$_{grad}$ &IMQ$_{conn}$\\ \hline
Mask RCNN~\cite{he2017mask} &44.3 M &18.37 &25.65 &0.45 &19.07 &24.22 &33.74 &2.27 &26.65\\
CascadePSP~\cite{cheng2020cascadepsp} &67.7 M &40.85 &51.64 &29.59 &43.37 &64.58 &74.66 &60.02 &67.20 \\
GCA~\cite{li2020natural} & 25.2 M &37.76 &51.56 &38.33 &39.90 &45.72 &61.40 &44.77 &48.81 \\
SIM~\cite{sun2021semantic} &  46.5 M &43.02 &52.90 &40.63 &44.29 &54.43 &66.67 &49.56 &58.12  \\
FBA~\cite{forte2020f} &  34.7 M &36.01 &51.44 &37.86 &38.81 &34.81 &48.32 &36.29 &37.23 \\
\gr{MGMatting}~\cite{yu2021mask} &\gr{+ 29.6 M} &\gr{51.67} &\gr{67.08} &\gr{53.03} &\gr{55.38} &\gr{57.98} &\gr{71.12} &\gr{66.53} &\gr{60.86}  \\
\gr{InstMatt}~\cite{sun2022human} &\gr{+ 29.7 M} &\gr{63.59} &\gr{78.14} &\gr{64.50} &\gr{67.71} &\gr{70.26} &\gr{81.34} &\gr{74.90} &\gr{72.60} \\ 
\hline
SAM~\cite{kirillov2023segment} &93.7 M &49.69 &61.44  &4.34 &51.84 &61.15 &74.01  &13.64 &65.85 \\ 
MAM &+ 2.7 M &54.15 &68.01 &30.47 &55.40 &68.78 &81.67  &51.79 &72.62 \\
\end{tabular}
\caption{Results on the instance image matting benchmark HIM2K. Metrics with \textit{mad}, \textit{mse}, \textit{grad}, and \textit{conn}  as subscript indicates the similarity metrics for IMQ are MAD, MSE, Gradient, and Connectivity, separately. $\uparrow$ means higher values indicate better performance for the IMQ metric. MAM shows clear improvements over SAM under different metrics with only 2.7M extra trainable parameters, much lighter compared to other mask-guided methods like MGMatting and InstMatt.} 
\vspace{-2mm}
\label{tab:him2k}
\end{table*}
\begin{table}[tb]
\centering
\begin{tabular}{c|cc}
Method &IMQ$_{mad}$$\uparrow$ &IMQ$_{mse}$$\uparrow$\\ \hline
Mask RCNN~\cite{he2017mask}  &20.26 &25.36 \\
CascadePSP~\cite{cheng2020cascadepsp}  &42.20 &52.91 \\
GCA~\cite{li2020natural} &33.87 &46.47  \\
SIM~\cite{sun2021semantic} &34.66 &46.60 \\
FBA~\cite{forte2020f}  &35.00 &47.54 \\
\gr{MGMatting}~\cite{yu2021mask} &\gr{30.64} &\gr{53.16} \\
\gr{InstMatt}~\cite{sun2022human} &\gr{51.10} &\gr{73.09} \\
\hline
SAM~\cite{kirillov2023segment} &49.87 &56.92 \\
MAM &54.40 &76.45  \\
\hline
\end{tabular}
\caption{Results on the instance matting benchmark RWP636.} 
\vspace{-4mm}
\label{tab:rwp}
\end{table}

We extensively evaluate the performance of MAM on six diverse image matting benchmarks. Through comprehensive evaluations using different metrics, we compare the performance of MAM with state-of-the-art image matting models on each benchmark. The results demonstrate that MAM consistently achieves comparable performance to specialized state-of-the-art models, reaffirming its versatility and effectiveness as a unified image matting solution.

\subsection{Implementation Details}

\noindent \textbf{Training Datasets} During the training process, we randomly select foreground instances from several image matting datasets, including Adobe Image Matting dataset~\cite{xu2017deep}, Distinctions-646~\cite{yu2020high}, AM2K~\cite{li2022bridging}, Human-2K~\cite{liu2021tripartite}, and RefMatte~\cite{li2023referring}, to ensure a diverse range of instance classes. For background images, we select them from two datasets: COCO~\cite{lin2014microsoft} and BG20K~\cite{li2022bridging} to provide a mix of both real-world and synthetic backgrounds.

\noindent \textbf{Evaluation Benchmarks} To evaluate the adaptive ability of MAM, we test it on a variety of image matting benchmarks including the semantic image matting benchmarks PPM-100~\cite{sun2021modnet}, AM2K~\cite{li2022bridging}, PM-10K~\cite{li2022bridging}, the instance image matting benchmark RWP636~\cite{yu2021mask}, HIM2K~\cite{sun2022human}, and the referring image matting benchmark RefMatte-RW100~\cite{li2023referring}. The box prompt is used for all benchmarks and the point prompt is only used in RefMatte-RW100. This comprehensive evaluation allows us to assess the generalization capability of MAM across various image matting tasks and benchmarks.

\noindent \textbf{Evaluation Metrics} We evaluate the accuracy of predicted alpha matte for MAM with commonly adopted evaluation metrics. Specifically, we employ Mean Absolute Difference (MAD), Sum of Absolute Difference (SAD), Mean Squared Error (MSE), Gradient (Grad), and Connectivity (Conn)~\cite{rhemann2009perceptually} as corresponding evaluation metrics. We scale MAD, MSE, Grad, and Conn by $10^3$, $10^3$, $10^{-3}$, and $10^{-3}$, respectively. Lower values indicate better performance for these metrics. Additionally, for instance-aware matting, we utilize Instance Matting Quality (IMQ)~\cite{sun2022human}, which takes recognition and matting accuracy into consideration simultaneously. Higher values indicate better performance for the IMQ metric.

\noindent \textbf{Experimental Settings} We trained MAM on a combination of training datasets using 8 RTX A6000 GPUs, with a batch size of 10 images per GPU. Each image was a combination of a randomly selected foreground instance and a background image. Images were cropped to a size of $1024 \times 1024$ and sent to a pre-trained ViT-B based SAM~\cite{kirillov2023segment} with a bounding box prompt of the target instance. The feature maps and masks output by SAM were then fed into the M2M module for alpha matte prediction. We employed the Adam optimizer with $\beta_1 = 0.5$ and $\beta_2 = 0.99$, trained for 20,000 iterations with warm-up for the first 4,000 iterations. The weight map $w_{os8}$ is always 1 at all pixels during training, while $w_{os4}$ changes to the mask guidance from SAM after the 4,000 iterations and $w_{os1}$ changes to the boundary of $\alpha_{os4}$ after the 4,000 iterations as well. We set 3 refinement blocks for the prediction of $\alpha_{os8}$, 3 refinement blocks for the prediction of $\alpha_{os4}$, and 2 refinement blocks for the prediction of $\alpha_{os1}$. As a result, the total trainable parameters of MAM is 2.7 million parameters. We applied cosine learning rate decay with an initial learning rate of 0.001 during training. During inference, we used a single GPU with a batch size of 1. Each image was resized to have its longer side at 1024 pixels and its shorter side was padded to 1024 pixels before being sent to MAM for alpha matte prediction of the target instance.

\begin{table}[tb]
\centering
\begin{tabular}{c|c|ccc}
Method &Prompt &SAD$_{all}$$\downarrow$ &MSE$_{all}$$\downarrow$ &MAD$_{all}$$\downarrow$\\ \hline
%MDETR-R101~\cite{kamath2021mdetr} &text  &131.58 &67.5 &75.1 \\
MDETR~\cite{kamath2021mdetr} &text  &131.58 &67.5 &75.1 \\
%CLIPSeg-ViT-B/16~\cite{luddecke2022image} &text &211.86 &117.8 &122.2 \\
CLIPSeg~\cite{luddecke2022image} &text &211.86 &117.8 &122.2 \\
\gr{CLIPMat}~\cite{li2023referring} &\gr{text} &\gr{107.81} &\gr{59.5} &\gr{62.0} \\
%\gr{CLIPMat-ViT-B/16-Refiner}~\cite{li2023referring} &\gr{text} &\gr{107.81} &\gr{59.5} &\gr{62.0} \\
%\gr{CLIPMat-ViT-L/16-Refiner}~\cite{li2023referring} &\gr{text} &\gr{85.83} &\gr{47.4} &\gr{49.5} \\
\hline
SAM~\cite{kirillov2023segment} &text &122.76 &67.9 &69.0  \\ 
MAM &text &120.10 &65.9 &67.5 \\ \hline
SAM~\cite{kirillov2023segment} &point &214.19 &123.8 &124.9  \\ 
MAM &point &168.82 &89.6 &97.7\\ \hline
SAM~\cite{kirillov2023segment} &box &33.51 & 17.9 &19.0  \\ 
MAM &box &29.24 &15.1 &16.6 \\
\hline
\end{tabular}
\caption{Results on the referring image matting benchmark RefMatte-RW100. MAM with box prompt can reach significantly better performance than with the text prompt.}
\vspace{-5mm}
\label{tab:refer_half}
\end{table}

\subsection{Main Results}
\noindent \textbf{Specialized vs Unified Model} We present a high-level comparison between specialized image matting models and MAM on the semantic, instance, and referring image matting benchmarks in Table~\ref{tab:specialvsmam}. It shows that MAM has clear improvements over SAM on all benchmarks. Furthermore, MAM shows comparable performance to each specialized image matting model and even reaches better performance on the HIM2K, RWP635, and RefMatte-RW100, which makes it a practical and feasible solution to unified image matting.

\noindent \textbf{Semantic Image Matting} We evaluate the performance of MAM on three semantic image matting benchmarks: PPM-100~\cite{sun2021modnet}, AM2K~\cite{li2022bridging}, and PM-10K~\cite{li2022bridging}, as presented in Table~\ref{tab:ppm} and Table~\ref{tab:am2k_pm10k}. The iterative refinement process is based on the $\alpha_{os8}$ prediction for all three benchmarks. On the PPM-100 benchmark, MAM achieves improvements of 6.2 MSE$_{all}$ and 3.9 MAD$_{all}$ over SAM. Similarly, on the AM2K benchmark, MAM outperforms SAM with enhancements of 7.64 SAD$_{all}$, 4.4 MSE$_{all}$, 4.5 MAD$_{all}$, 41.54 Grad$_{all}$, and 7.59 SAD$_{tri}$. 
%Notably, MAM exhibits superior performance while being more lightweight compared to other models on the AM2K benchmark. On the PM-10K benchmark, MAM achieves improvements of 17.45 SAD$_{all}$, 19.4 MSE$_{all}$, 9.9 MAD$_{all}$, 10.47 Grad$_{all}$, and 7.96 SAD$_{tri}$ over SAM, respectively.

\begin{table}[tb]
\centering
\begin{tabular}{c|c|cc}
&Model &\multicolumn{2}{c}{Natural Subset}\\
Method  & Size  &IMQ$_{mad}$ &IMQ$_{mse}$ \\ \hline
SAM~\cite{kirillov2023segment} &93.7 M &50.47 &61.66  \\
+ Mask-Select &93.7 M &61.15 &74.01  \\  \hline
MAM Baseline & 1.0 M  &52.82 &71.82  \\
+ Multi-Scale Prediction & 2.7 M  &60.11 &74.74  \\
+ Iterative Refinement & 2.7 M  &65.44 &78.93\\
+ Multi-Dataset Training & 2.7 M &68.37 &81.56 \\
\hline
\end{tabular}
\caption{Ablation study of MAM on the HIM2K benchmark. The MAM Baseline is built upon the SAM model with the box prompt. The other strategies are gradually added to the MAM Baseline and end up with 2.7 M extra trainable parameters.}
\label{tab:ablation_half}
\vspace{-3mm}
\end{table}
\noindent \textbf{Instance Image Matting} In Table~\ref{tab:him2k} and Table~\ref{tab:rwp}, We evaluate MAM on two instance image matting benchmarks: HIM2K~\cite{sun2022human} and RWP636~\cite{yu2021mask}. For HIM2K, the iterative refinement is based on prediction mask $m$ since it contains multiple instances per image and starting from $m$ removes false positive predictions. 
Compared to other state-of-the-art methods on HIM2K, MAM reaches comparable performance with only 2.7 M extra trainable parameters, which is only 10\% of the specialized models like MGMatting and InstMatt, which use the mask guidance from Mask RCNN. On the RWP636 benchmark, we apply the iterative refinement from $\alpha_{os8}$ and MAM reaches the new state-of-the-art with 54.40 IMQ$_{mad}$  and 76.45  IMQ$_{mse}$.

%On the HIM2K synthetic subset, SAM sets a baseline of 49.69 IMQ$_{mad}$, 61.44 IMQ$_{mse}$, 22.22 IMQ$_{grad}$, 33.33 IMQ$_{conn}$ while MAM reaches 53.66 IMQ$_{mad}$, 68.74 IMQ$_{mse}$, 32.80 IMQ$_{grad}$ 55.23 IMQ$_{conn}$, which has clear improvements over SAM. Then, we further evaluate SAM and MAM on the natural subset of HIM2K. MAM further make improvements of 7.32 IMQ$_{mad}$, 7.55 IMQ$_{mse}$, 42.40 IMQ$_{grad}$, 6.32 IMQ$_{conn}$ over SAM. Compared to other state-of-the-art methods on HIM2K, MAM reaches comparable performance with only 2.7 M trainable parameters, which is only 10\% of the specialized InstMatt. On the RWP636 benchmark, SAM sets a strong baseline of 49.88 IMQ$_{mad}$ and 56.94 IMQ$_{mse}$. Then, we apply the iterative refinement from $\alpha_{os8}$ and MAM reaches the new state-of-the-art with 54.40 IMQ$_{mad}$  and 76.45  IMQ$_{mse}$.

\noindent \textbf{Referring Image Matting} In Table~\ref{tab:refer_half}, we present the evaluation of MAM on the RefMatte-RW100 benchmark~\cite{li2023referring}, a recently introduced referring image matting benchmark. While previous methods rely on text prompts for referring image matting, we leverage the bounding boxes and text descriptions as the prompts for SAM. Considering the text prompt for SAM has not been released yet, we use Grounded-SAM~\cite{liu2023grounding} to support text prompt guidance. Remarkably, MAM achieves superior performance when utilizing the bounding box as the prompt for SAM, surpassing the text-based methods CLIPSeg and CLIPMat by a significant margin. Moreover, the use of bounding boxes as prompts offers user-friendly and intuitive interactions, as users find it easier to provide bounding boxes compared to composing a fixed text paragraph. This observation suggests that the bounding box prompt is more effective for interactive image matting than the text or point prompt for referring image matting.

\subsection{Ablation Study}
We conduct comprehensive ablation studies on the M2M module of MAM, considering that SAM remains frozen during the training process. To assess the performance of MAM, we select the real-world subset of the HIM2K benchmark.

\begin{figure}[tb]
\centering
\includegraphics[width=0.5\textwidth]{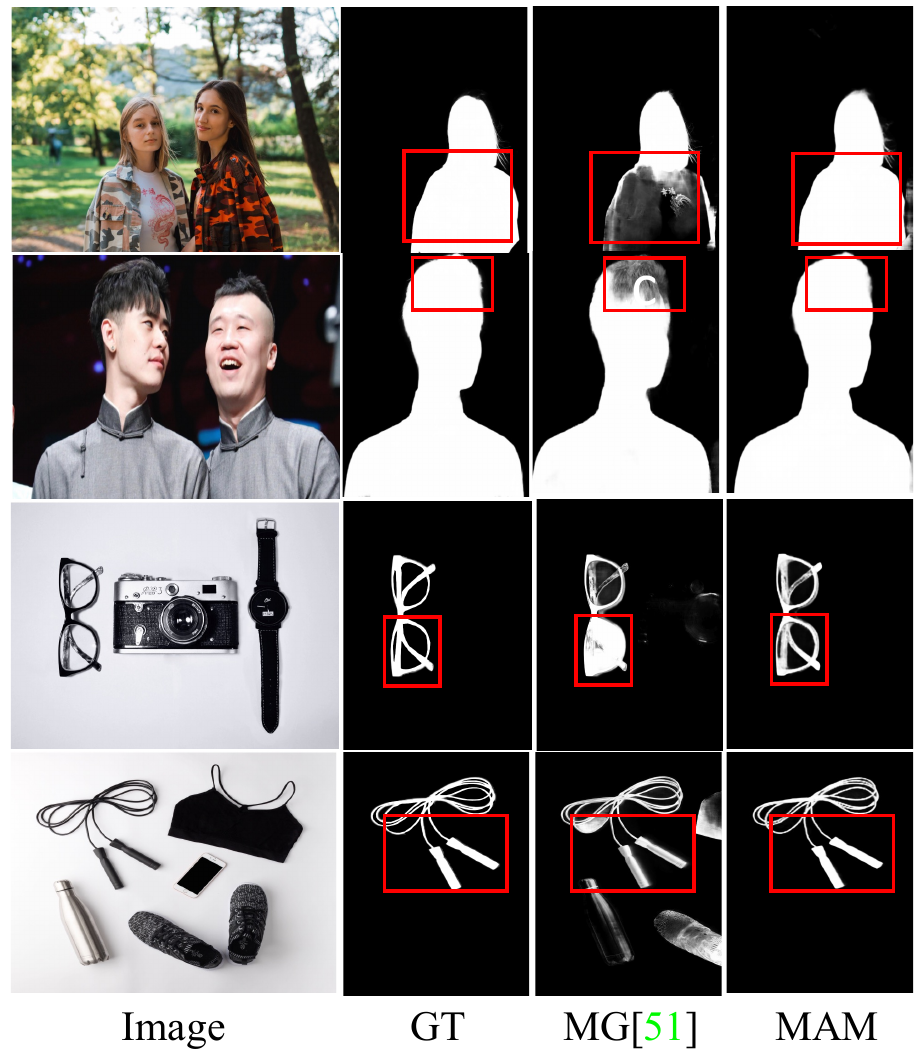}
\caption{Visualizations of alpha matte predictions from MGMatting and MAM. Improvements are highlighted in the red boxes.}
\label{fig:vis_matt}
\vspace{-4mm}
\end{figure}

\begin{figure*}[tb]
\centering
\includegraphics[width=1.0\textwidth]{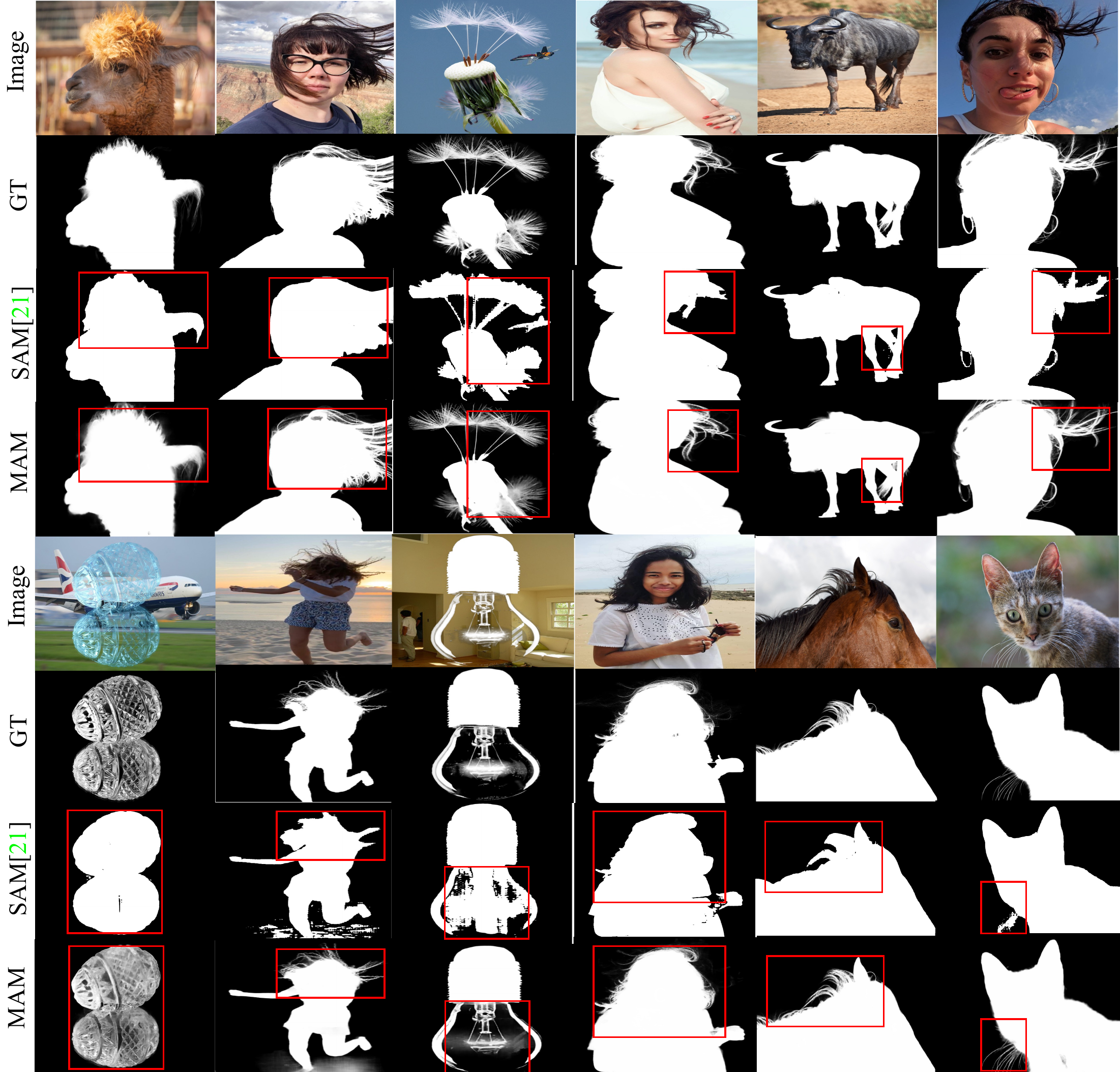}
\caption{Visualizations of mask and alpha matte predictions from SAM and MAM. Improvements are highlighted in the red boxes.}
\label{fig:vis}
\vspace{-3mm}
\end{figure*}

\noindent \textbf{SAM on HIM2K} We begin by evaluating the pre-trained ViT-B-based SAM using bounding boxes and points as prompts for the target instance. SAM with box-based prompts significantly outperforms the point-based prompts and the final mask output is selected based on the mask with the highest Intersection over Union (IoU) score with the bounding box. SAM demonstrates strong performance on the HIM2K benchmark, achieving 61.15 IMQ$_{mad}$ and 74.01 IMQ$_{mse}$ on the natural subset.

\noindent \textbf{Building MAM} We then construct the M2M baseline by integrating the M2M module, which takes SAM's mask and feature maps, as well as the image, as inputs. This baseline, comprising 3 connected refinement blocks and predicting at 1/16 resolution, yields inferior performance compared to SAM, as the low-resolution predictions lack fine details of the alpha matte. However, by gradually incorporating multi-scale predictions and iterative refinement, as described in Section~\ref{m2m}, the performance of MAM improves. Additionally, the adoption of multi-dataset training, as outlined in Section~\ref{mam}, further enhances performance, resulting in 68.37 IMQ$_{mad}$ and 81.56 IMQ$_{mse}$ on the natural subset. Subsequently, we assess MAM's performance on other benchmarks without retraining to validate its generality and adaptability.

\subsection{Visualization}
In Figure~\ref{fig:vis_matt}, we compare matting performance between MGMatting and MAM of images that contain multiple instances. They both leverage mask guidance from SAM. It shows that MAM is able to give more accurate alpha matte predictions with only 10\% parameters compared to MGMatting under the same mask guidance. It also has fewer false positive predictions in other instances. In Figure~\ref{fig:vis}, we further provide visualizations of the mask and alpha matte predictions from SAM and MAM. These images are selected from the semantic image matting benchmarks and contain a single instance that can be a person, animal, or transparent object. The visualizations demonstrate that MAM achieves significantly improved predictions in the transition areas without the trimap guidance, which highlights the superior performance of MAM in refining and enhancing the quality of alpha matte predictions.
%Additionally, MAM effectively addresses some of the holes present in the mask predictions generated by SAM. These visual comparisons highlight the superior performance of MAM in refining and enhancing the quality of alpha matte predictions.

%\subsection{Limitation}
%One limitation of MAM is its dependence on the feature maps and mask predictions generated by SAM. If SAM produces incorrect mask predictions, such as returning the mask of an irrelevant instance like the bottom left case in Figure~\ref{fig:vis}, MAM faces challenges in rectifying these errors. The iterative refinement process may inadvertently propagate such errors, leading to inaccurate final predictions. Therefore, addressing the issue of enabling MAM to effectively correct instance-level prediction errors remains an open question that requires further investigation.

\section{Conclusion}
In this paper, we introduce Matting Anything Model (MAM), which uses the Segment Anything Model (SAM) as a guidance module with a lightweight Mask-to-Matte (M2M) module to refine the mask output into the alpha matte of the target instance. M2M is designed to handle various image matting tasks, including semantic, instance, and referring image matting, using a single model based on user prompts including points, boxes, and text. We evaluate MAM on six image matting benchmarks and demonstrate that it achieves comparable performance to the specialized state-of-the-art methods under various evaluation metrics. Our proposed model offers a more versatile and efficient solution for interactive and unified image matting. 

{\small
\bibliographystyle{ieee_fullname}
\bibliography{references}
}
\end{document}